# Underwater Waste Detection Using Deep Learning: A Performance Comparison of YOLOv7–10 and Faster R-CNN

**UMMPK Nawarathne[1#], HMNS Kumari[2], and HMLS Kumari[3]**
[1] Faculty of Computing, Sri Lanka Institute of Information Technology, Sri Lanka
[2] Faculty of Information Technology and Communication Sciences, Tampere University, Finland
[3] Computing Centre, Faculty of Engineering, University of Peradeniya, Sri Lanka

[#]malithi.n@sliit.lk

**ABSTRACT** Underwater pollution is one of today's most significant environmental concerns, with vast volumes of garbage found in seas, rivers, and landscapes around the world. Accurate detection of these waste materials is crucial for successful waste management, environmental monitoring, and mitigation strategies. In this study, we investigated the performance of five cutting-edge object recognition algorithms, namely YOLO (You Only Look Once) models, including YOLOv7, YOLOv8, YOLOv9, YOLOv10, and Faster Region-Convolutional Neural Network (R-CNN), to identify which model was most effective at recognizing materials in underwater situations. The models were thoroughly trained and tested on a large dataset containing fifteen different classes under diverse conditions, such as low visibility and variable depths. From the above-mentioned models, YOLOv8 outperformed the others, with a mean Average Precision (mAP) of 80.9%, indicating a significant performance. This increased performance is attributed to YOLOv8's architecture, which incorporates advanced features such as improved anchor-free mechanisms and self-supervised learning, allowing for more precise and efficient recognition of items in a variety of settings. These findings highlight the YOLOv8 model's potential as an effective tool in the global fight against pollution, improving both the detection capabilities and scalability of underwater cleanup operations.

**INDEX TERMS** Faster R-CNN, Object detection, Underwater garbage detection, YOLOv8

## I. INTRODUCTION

Maintaining a safe and sanitary urban environment is vital as the global population and industries continue to increase. Illegal garbage disposal harms the ecology and poses a health risk. If not addressed promptly, it can lead to serious health issues and environmental degradation [1], [2]. Littering the surroundings and contaminating the environment harms many people, including children, the elderly, travellers, small and large companies, as well as animals. Hence, garbage management is a global problem that has been a concerning topic throughout history.

Underwater pollution, particularly from plastics and chemical contaminants, has serious effects on marine biodiversity and ecosystems [1]. The control of underwater pollution is vital for sustaining marine ecosystems, protecting biodiversity, and ensuring the health of human populations who rely on these settings. Underwater pollution, primarily from plastics, endangers marine life, killing about 100 million species each year [1]. This pollution also disturbs the food chain since microplastics are consumed by small marine species and accumulate in larger predators, eventually harming human food supplies [3]. Furthermore, underwater pollution damages coral reefs, which are critical for supporting marine biodiversity and protecting coasts from erosion and storm surges [4].

The economic consequences of underwater pollution are significant, endangering industries such as tourism, fishing, and shipping. S. C. Gall and R. C. Thompson [5] calculated that marine debris costs the Asia-Pacific area approximately $1.26 billion per year. Therefore, underwater pollution has far-reaching economic consequences, harming industries including tourism, fishing, and coastal communities. Furthermore, contaminants such as heavy metals and chemicals that settle in the water can have long-term detrimental impacts on marine and human life, resulting in chronic diseases and ecosystem imbalances [6]. Marine debris, especially plastics, has been demonstrated to harm fishing gear, impair catch quality, and degrade the aesthetic value of coastal areas, resulting in substantial financial losses. According to research by S.Newman et al. [7], the worldwide cost of marine plastic pollution is projected to be billions of dollars per year due to its effects on tourism, fisheries, and marine ecosystems.

Addressing underwater pollution requires a multifaceted approach that includes sophisticated research, robust legislation, and the development of novel technology to manage and mitigate its effects properly. Machine learning, Artificial Intelligence (AI), and autonomous systems are examples of recent technological breakthroughs that provide promising solutions for detecting, classifying, and removing underwater contaminants. For example, AI-powered image recognition systems, such as convolutional neural networks (CNNs), can dramatically enhance the identification of various forms of garbage, allowing for more precise and efficient cleanup operations.

Hence, this study aims to conduct a comparative evaluation of state-of-the-art object detection models, YOLOv7 through YOLOv10, and Faster R-CNN, in the context of underwater garbage detection. By analysing the performance of these models on a 15-class underwater waste materials dataset, this study tries to determine which model operates most successfully in identifying waste materials under challenging marine conditions.

## II. LITERATURE REVIEW

A.R.Faijunnahar et al. [8] used CNN and Support Vector Machine (SVM) algorithms to detect plastic waste from the garbage. The CNN model achieved a training accuracy of 0.95, whereas the SVM model obtained a training accuracy of 0.71. The testing accuracy was 0.92 for the CNN model and 0.62 for the SVM model. However, by relying solely on basic CNN and SVM, this study was constrained in its ability to explore more advanced architectures under different varying conditions.

A study conducted by Q. Chen and Q. Xiong [9] examined the use of an upgraded YOLOv4 model for garbage classification detection in order to increase the accuracy and efficiency of waste sorting systems. The authors proposed modifications to the conventional YOLOv4 architecture to effectively handle the various and complicated waste materials, which can vary greatly in size, shape, and colour. These enhancements included improving the network structure for faster processing and using advanced feature extraction techniques to capture the distinguishing properties of various forms of garbage. The upgraded model was trained and evaluated on a dataset containing diverse types of rubbish, displaying superior performance in reliably recognizing and categorizing garbage when compared to the original YOLOv4 and other traditional models. Although the improved YOLOv4 model performed well, it was trained using three types of garbage categories, thereby limiting the model's generalizability to more complex and diverse datasets.

O. Goxha et al. [10] investigated the application of machine learning algorithms in underwater image identification, with a special emphasis on their usage for cleaning. Poor visibility, light scattering, and the presence of particulate matter, each of which can reduce image quality, provide distinct challenges for underwater image detection. The research examined various machine learning algorithms for improving picture detection performance in these demanding situations. CNNs, SVMs, and deep learning models were recognized for their ability to efficiently process and evaluate underwater images. This study highlighted the need for preprocessing measures such as picture enhancement and noise reduction, as well as the use of color and texture-based features, which are critical for increasing the clarity and accuracy of recognized objects.

X.Teng et al. [11] presented an improved YOLOv5 model designed for detecting underwater garbage, specifically targeting plastic waste. The model's enhancements included re-clustering anchor boxes using the KMeans++ algorithm and optimizing the loss function, which resulted in an impressive detection accuracy of 88.7% and a mean average precision (mAP) of 90.6%. These results indicated a significant improvement over previous models, highlighting the potential of the improved YOLOv5 for practical use in autonomous underwater vehicles (AUVs) for garbage detection and collection.

The work conducted by Z. Hu and C. Z. Xu [12] described an upgraded identification technique for underwater plastic garbage based on an improved version of the YOLOv5n model. The issues addressed included the diverse underwater environment and insufficient lighting, which often restrict detection accuracy and raise processing needs. The authors introduced significant improvements to the YOLOv5n model, including a CB2D module in the backbone network, which reduced model parameter size while enhancing detection accuracy. They also incorporated the ConvNext-Block (CNeB) module in the feature pyramid to improve small object detection, and the alpha-Intersection over Union (IoU) loss function for enhanced bounding box regression. The experimental findings demonstrated significant improvements, with the Average Precision-50 (AP50) value and overall Average Precision (AP) recorded at 12.25% and 13.69%, respectively, compared to the original YoloV5n model. However, by limiting this study to a single YOLO5n architecture without conducting a comprehensive comparison against other advanced models, the reliability of the results in different underwater conditions remains uncertain.

F. Zhang et al. [13] proposed an underwater object detection algorithm based on an improved version of the YOLOv8 model named YOLOv8-CPG. The authors have reported that Precision and Recall show improvements of 2.76% and 2.06%. Additionally, mAP50 and mAP50-95 metrics have increased by 1.05% and 3.55% [14]. This study is another example that depicts the prowess of YOLOv8 model. However, similar to the study conducted by Z. Hu and C. Z. Xu [12], the authors limited their work to improving a single YOLO architecture without addressing broader challenges in varying underwater conditions and leaving the relative strengths of different approaches for underwater object detection unexplored.

A. Guo et al. [14] conducted a study that optimizes YOLOv8s for an underwater target detection method. Experiments have been conducted on three underwater datasets, RUOD, UTDAC2020, and URPC2022, and the authors have obtained results of mAP50 of 86.8%, 84.3%, and 84.7% per respective dataset, which proves the efficiency of the YOLOv8 model. The experimental results demonstrate that the proposed UW-YOLOv8 significantly reduces both the model size and processing demands, satisfying the criteria for real-time detection. However, the authors acknowledged the poor performance of this model under crowded and low-visibility environments, highlighting the need for further improvements.

While various studies have been conducted to improve individual object identification architectures for underwater environments, there is still a lack of a comprehensive comparison of the cutting-edge models specifically applied to underwater garbage detection. Existing research generally concentrates on model enhancements or optimisations for single architectures such as YOLOv5 or YOLOv8, but few studies have systematically compared the performance of newer YOLO variants with well-established methods such as Faster R-CNN in this domain. However, these types of comparative assessments are crucial to determining the most effective and efficient models for real-world underwater garbage detection applications, given the rapidly changing nature of object detection technologies. Therefore, this study aims to fill this gap by providing a thorough assessment of these models on a challenging multi-class underwater garbage dataset, providing insightful information to direct future studies and implementation in marine environmental monitoring.

## III. METHODOLOGY
### A. Data

The data used for the analysis of this study were obtained from an online data repository [15]. These data contained images of the undersea, which were spoiled by rubbish and abandoned trash. Initially, there were three different datasets, namely, the training, test, and validation sets. The training dataset contained 3628 images, the test dataset had 501 images, and the validation dataset included 1001 images along with their respective labels. Each of these datasets consisted of fifteen classes of garbage, which are of the following types: masks, cans, cell phones, electronics, glass bottles, gloves, metal, net, polythene bags, plastic bottles, rods, sunglasses, tires, and miscellaneous. Figures 1 and 2 depict some of the images contained in the original dataset.

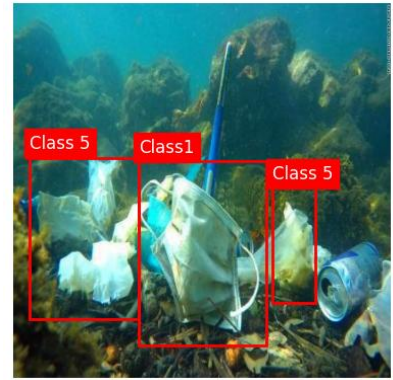

Figure 1. Images of the undersea trash contained in the original dataset [15]

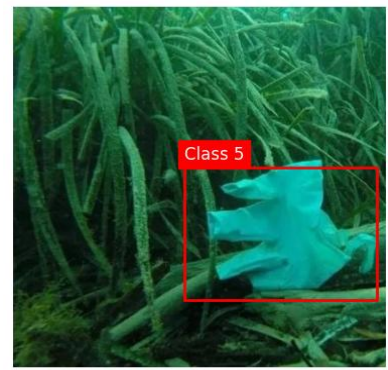

Figure 2. An image of the undersea trash contained in the original dataset [15]

These data were then trained using five different object detection models, named YOLOv7, YOLOv8, YOLOv9, YOLOv10, and Faster R-CNN. However, prior to discussing these models in detail, it is essential to present the experimental pipeline of this study. Therefore, Figure 3 illustrates a general overview of the workflow, briefly detailing each step in order to provide a better understanding of this study's structure, including the methodology carried out.

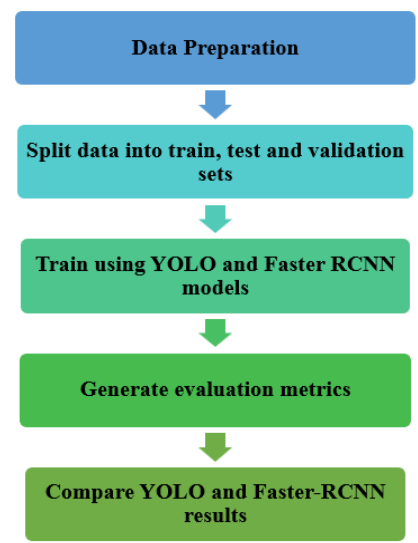

Figure 3. A general overview of the workflow

## B. Object Detection Models Used to Train the Data

YOLO is a well-known object detection framework that has seen significant evolution from its introduction to YOLOv10, with each version offering improvements in speed, accuracy, and model design.

*1) YOLOv7*: The YOLOv7, developed by Wang et al. [16] represents a significant advancement in real-time object identification, with a focus on striking a balance between accuracy and computing efficiency. It introduced a revised architecture with enlarged convolutions and optimized anchor settings, resulting in shorter inference times while maintaining good detection accuracy. The addition of Cross Stage Partial Connections (CSP) and Path Aggregation Network (PANet) improved YOLOv7's efficiency for edge devices with low resources [16].

*2) YOLOv8:* The YOLOv8, which was introduced by G. Jocher, A. Chaurasia, and J. Qi [17], improved upon YOLOv7 by focusing on increasing detection accuracy, particularly for smaller objects and complex situations. YOLOv8's main breakthrough was the incorporation of transformer-based modules, which increased the model's capacity to capture long-range dependencies in images. This version also included advanced augmentation techniques during training, such as mosaic and mixup, which improved generalization and resilience across several datasets. YOLOv8 gained particular significance in applications that require higher precision, such as medical imaging and autonomous driving. The YOLOv8 architecture, which was used for this study, is depicted in Figure 4 [18].

*3) YOLOv9:* The YOLOv9, developed by J. Redmon et al. [19], introduced hybrid models that combined CNNs and vision transformers (ViTs), resulting in a higher performance on various benchmarks. YOLOv9's architecture was designed to minimize the trade-offs between speed and accuracy by including dynamic head networks and adaptive anchor-free methods, making it highly flexible for varying object scales and aspect ratios. This version gained popularity in settings that require both real-time processing and great precision, such as video surveillance and industrial automation.

*4) YOLOv10:* YOLOv10 is one of the most recent innovations, pushing the limits of object detection with a unified architecture that includes self-supervised learning approaches invented by A. Wang et al. [20]. YOLOv10's most notable feature is its capacity to learn from unlabelled data, which significantly reduces the need for complex labelled datasets. Furthermore, it employs an enhanced version of the attention process found in transformers, resulting in remarkable precision, particularly in cluttered or obscured scenes. This tool is critical for applications such as autonomous drones and complicated robotic vision systems, where labelling data is difficult or expensive.

*5) Faster R-CNN:* The Faster R-CNN is a significant advancement in the object identification field, expanding on the capabilities of prior R-CNN models using a Region Proposal Network (RPN) to enhance detection speed and accuracy. Faster R-CNN was developed by S.Ren et al. [21] in 2015, and is intended to solve inefficiencies in its predecessors, specifically the slow region proposal algorithms used in R-CNN and Fast R-CNN. The RPN, a fully convolutional network, is the central innovation of Faster R-CNN. It quickly generates high-quality region recommendations by sharing convolutional layers with the detection network, dramatically decreasing computational overhead and enabling end-to-end training.

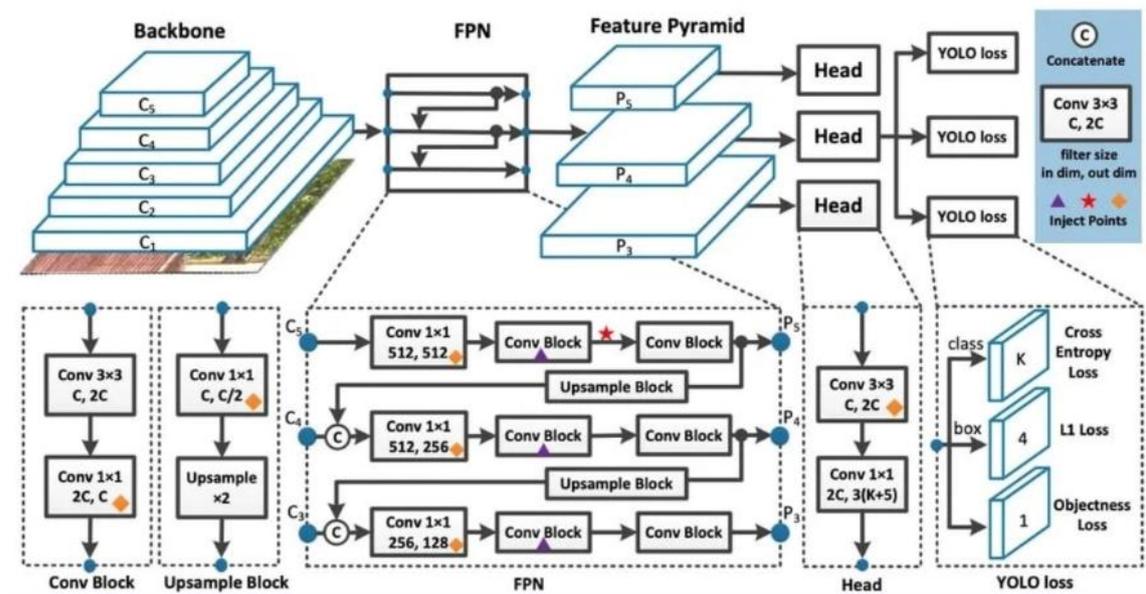

Figure 4. YOLOv8 model architecture [18]

However, before applying any YOLO object detection models to these images, it is crucial to prepare the data. Therefore, distinct folders for train, test, and validation images were created as train, test, and val since this is the name order accepted by the YOLO models. For the training data with Faster R-CNN, the model was fed with data in the traditional way. The models were then trained with 50 epochs, considering a batch size of 64, where the image size was 128. After training the data with the five different object detection models, the performance of the models was evaluated using precision, recall, and mean average precision (mAP), which are defined in the YOLO documentation.

*C. Evaluation Metrics*

*1) Precision:* Precision measures the proportion of genuine positives out of all positive predictions, indicating the model's ability to prevent false positives [22].

*2) Recall:* On the other hand, recall estimates the proportion of true positives out of all actual positives, indicating the model's capacity to recognize all instances of a class [23].

*3) Mean Average Precision (mAP):* mAP broadens the definition of average precision (AP) by determining the average AP value across various object classes. This is useful in multi-class object identification scenarios to provide a thorough evaluation of the model's performance [22].

**IV. RESULTS AND DISCUSSION**

This study used 5130 images of fifteen types of garbage. Figure 5 depicts how the data have been distributed among these fifteen classes.

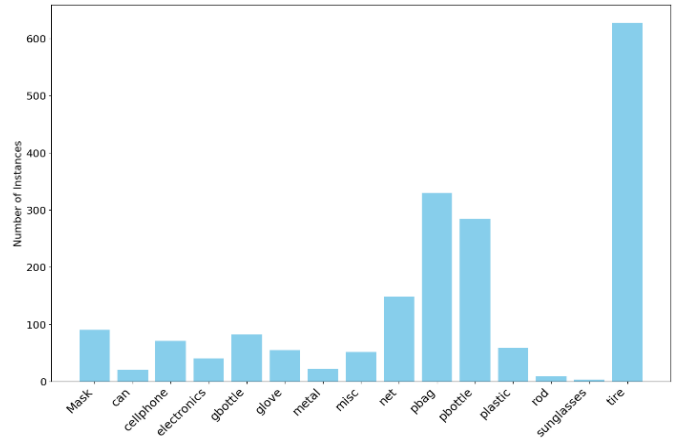

Figure 5. Distribution of images among the classes

After training the dataset with the respective object detection algorithms, the evaluation metrics were obtained, as shown in Table 1.

Table 1. Performance of the object detection models

| Model | Precision | Recall | mAP |
|---|---|---|---|
| YOLOv7 | 0.663 | 0.656 | 0.679 |
| YOLOv8 | 0.832 | 0.727 | 0.809 |
| YOLOv9 | 0.789 | 0.729 | 0.79 |
| YOLOv10 | 0.688 | 0.686 | 0.716 |
| Faster R-CNN | 0.484 | 0.629 | 0.615 |

According to Table 1, it is evident that the YOLOv8 model has performed well considering the precision, recall, and mean average precision values. However, Faster R-CNN has the lowest evaluation metric values compared to YOLO models. It is crucial to evaluate the confusion matrix of the YOLOv8 model. Therefore, the confusion matrix of YOLOv8 was obtained as in Figure 6, and it illustrates that the YOLOv8 model has performed well.

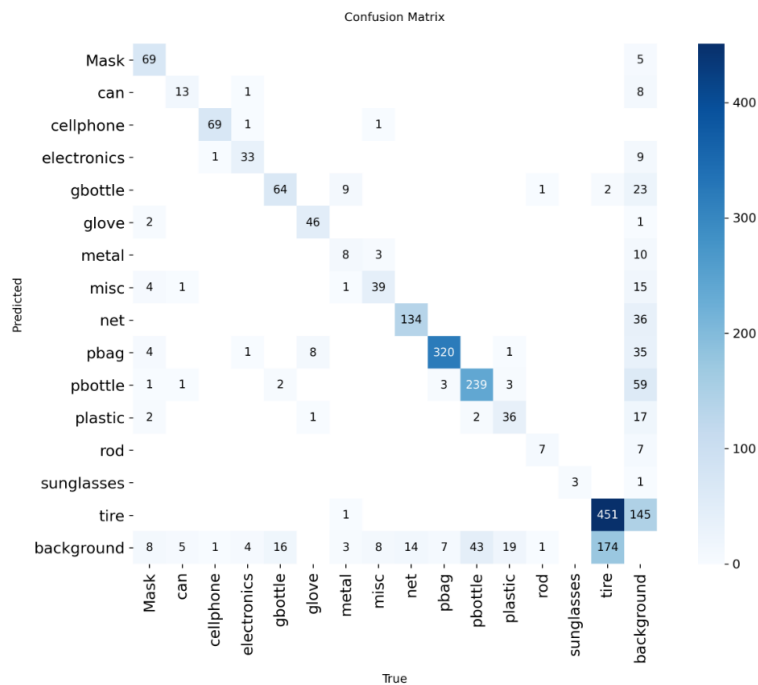

Figure 6. Confusion matrix of the resulting YOLOv8 model

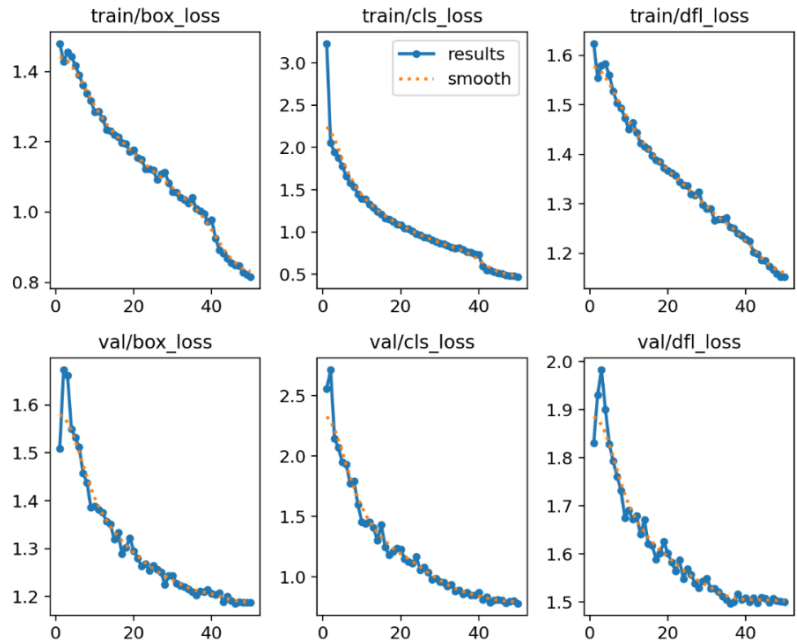

Figure 7. Metric curves of the resulting YOLOv8 model

In addition, metric curves for the YOLOv8 model were obtained, as shown in Figure 7. These curves indicate that the YOLOv8 model is progressing well, with loss values decreasing and recall and mAP metrics improving.

Furthermore, model predictions were obtained for the YOLOv8 model, and a few samples of these predictions are depicted in Figures 8,9,10, and 11.

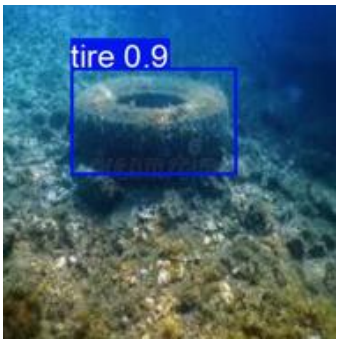

Figure 8. Predictions of the resulting YOLOv8 model

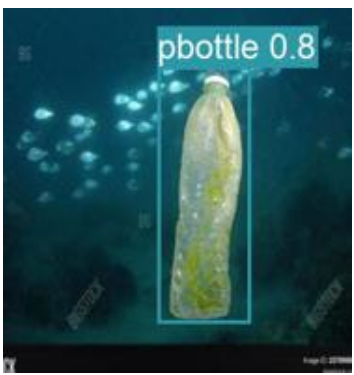

Figure 9. Predictions of the resulting YOLOv8 model

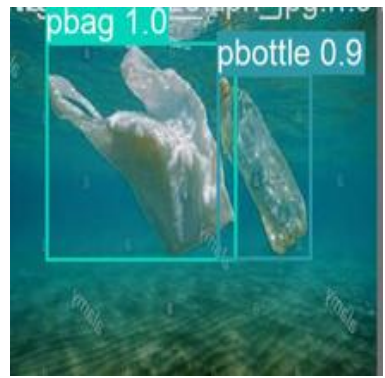

Figure 10. Predictions of the resulting YOLOv8 model

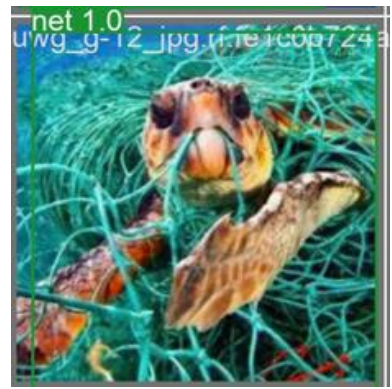

Figure 11. Predictions of the resulting YOLOv8 model

According to the model predictions shown in Figures 8,9,10, and 11, the YOLOv8 model has been able to detect various items precisely. Hence, it is evident that YOLOv8 has the ability to detect garbage materials efficiently.

Despite the promising results, this study has several limitations. Firstly, the dataset used in this study does not fully capture the real-world undersea pollution and other environmental conditions, including low visibility, varying illumination, and cluttered backgrounds. In addition, the models were evaluated on a smaller number of classes, which limits the generalizability of the results when applying to different waste types. For future work, this work can be expanded using a larger dataset collected from multiple locations and categories. Moreover, integrating domain adaptation and transfer learning techniques may help the models generalize better across different underwater settings. Furthermore, combining real-time detection capabilities with underwater robotic systems could have a tremendous impact on environmental monitoring and cleanup activities.

## V. CONCLUSION

This study used a dataset of 5130 images across fifteen garbage categories to demonstrate recent advances in deep learning techniques for underwater rubbish detection using several object detection models. This employed YOLOv7, YOLOv8, YOLOv9, YOLOv10, and Faster R-CNN on the dataset and identified that YOLOv8 surpassed its other models as well as Faster R-CNN, with a mean average precision of 80.9%. The metric curves and predictions obtained for this model proved its ability to detect the. objects with a higher accuracy rate. Finally, this study concludes that YOLOv8 performs reliably and effectively in detecting a variety of underwater waste categories.